\documentclass[runningheads]{llncs}

\usepackage{graphicx}
\usepackage{amsmath}
\usepackage{amssymb}
\usepackage{booktabs}
\usepackage{multirow}
\usepackage{float}
\usepackage{color}
\usepackage[hidelinks,hypertexnames=false]{hyperref}
\usepackage{algorithm}
\usepackage{algorithmic}
\usepackage{longtable,array}
\usepackage{pdflscape}
\usepackage{multicol}

\begin{document}

\title{Simplicity Goes Far: Auditing Prompt Optimizers with Demonstration-Conditioned Prompt Search}
\titlerunning{Auditing Prompt Optimizers with DCPS}

\author{Bohe Chen\inst{1,2}}
\authorrunning{B. Chen}
\institute{Beijing University of Posts and Telecommunications, Beijing, China
\and Queen Mary University of London, London, UK\\
\email{bhchen@bupt.edu.cn}}

\maketitle

\begin{abstract}
Automatic prompt optimization is growing more elaborate. Recent methods
stack policy-gradient updates, trace reflection, and evolutionary search. These
methods share a generate-and-select step; without controlling for it, reported
gains may conflate the contributions of optimizer mechanisms and larger
sampling budgets. We isolate this step as
\textbf{D}emonstration-\textbf{C}onditioned \textbf{P}rompt \textbf{S}earch
(DCPS), repurposing the Automatic Prompt Engineer (APE) recipe as a
mechanism-attribution control: it freezes the generator, conditions it on
randomly sampled demonstrations, scores candidates on validation data, and
keeps the best---no learned update, reflection, or evolution. Across two
optimizer families, DCPS replaces Proximal Policy Optimization (PPO) in StablePrompt, matching classification
accuracy within $0.5$~pp at roughly one-third the measured cost; in DSPy-based
compound AI systems, under the reported budgets, it matches GEPA's macro
accuracy on GPT-4.1-mini while using $55\%$ fewer rollouts at $32\%$ lower
measured API cost on average. The audit reveals a task-dependent pattern:
proposal and selection recover much of heavier optimizers' performance on
structured and instruction-following tasks, while additional mechanisms retain
advantages in some open-ended and reasoning-heavy settings. We recommend
benchmarking prompt optimizers against generate-and-select controls and
reporting rollout counts, token usage, and dollar costs alongside accuracy. Code and artifacts: \url{https://github.com/bhc6/DCPS}.

\end{abstract}

\section{Introduction}

Automatic prompt optimization (APO) has become increasingly elaborate. Reinforcement learning (RL) methods~\cite{deng2022rlprompt,kwon2024stableprompt} often rely on long rollout streams and policy-optimization pipelines, while reflection-based methods~\cite{gepa} use natural-language feedback to diagnose failures and rewrite prompts. These mechanisms can improve peak accuracy, but they also increase compute and API costs and make attribution harder. When a prompt optimizer works, is the gain coming from policy updates, reflection, and Pareto-style evolution, or from a simpler process of proposing prompts and selecting the best one on validation data?

To answer this question, we study \textbf{DCPS (Demonstration-Conditioned Prompt Search)}, a deliberately minimal control built from the Automatic Prompt Engineer (APE) generate-and-select pattern~\cite{zhou2023ape}. Unlike APE, which introduced this pattern as a standalone prompt-search algorithm, DCPS repurposes it as a control for mechanism attribution: the same generate-and-select loop is inserted into existing optimizer settings to probe what policy updates, reflection, merging, or Bayesian search add beyond validation-based selection. Demonstrations condition the generator to propose prompts but are not automatically carried into the final target-model prompt; DCPS is therefore neither a few-shot example selector nor a new optimizer, but a diagnostic tool whose contribution lies in the audit conclusions it enables.

We apply this control in two representative APO settings, differing only in the selection rule. \textbf{StablePrompt-DCPS} instantiates DCPS inside StablePrompt~\cite{kwon2024stableprompt}, freezing the generator and removing Proximal Policy Optimization (PPO) updates while preserving demonstration-conditioned proposal and Top-$K$ selection. \textbf{DCPS-Compound} instantiates the same control on GEPA-style compound AI systems~\cite{gepa}, keeping the task and rollout protocol but removing trace reflection, prompt mutation, merging, and Pareto evolution. We further report the Relative Cost-Efficiency Index (RCEI) as an auxiliary view of the token--performance trade-off.

\paragraph{Contributions.}
\begin{enumerate}
    \item \textbf{A reusable audit methodology.} We formalize DCPS as a four-stage control template and instantiate it inside two structurally different optimizer families---an RL-based pipeline (StablePrompt) and compound AI systems (GEPA and the heavy-budget MIPROv2 configuration)---to probe how much individual optimizer mechanisms add beyond validation-based selection.
    \item \textbf{A diagnostic boundary for optimizer complexity.} Across both settings, proposal and selection recover most gains on structured tasks at lower cost, while policy updates and reflection retain advantages on some open-ended and reasoning-heavy tasks. The observed pattern identifies where optimizer overhead may be justified without treating task category as a deterministic boundary.
    \item \textbf{Cost-aware evaluation.} We report accuracy jointly with rollout counts, token usage, and dollar costs, and use RCEI as an auxiliary cost-efficiency view, giving a replicable protocol for comparing optimizers under their actual budgets.
\end{enumerate}

\section{Related Work}
\label{sec:related-work}

\textbf{Generate-and-select controls.} Instruction Induction~\cite{honovich2022instruction} and APE~\cite{zhou2023ape} elicit natural-language instructions from demonstrations and select them by task performance. Notably, APE itself reports that its optional iterative (Monte~Carlo) search saturates within a few rounds and is disabled by default---an early hint that added search contributes little once proposal and selection are in place. DCPS inherits this structure but changes its role: rather than a standalone optimizer, it is a controlled ablation for existing pipelines, systematizing that hint across policy updates, reflection, and evolution.

\textbf{RL updates and reflection loops.} Existing APO methods~\cite{ramnath2025systematic} split into two paradigms. RL-based optimizers train an auxiliary generator to reshape discrete token distributions (RLPrompt, StablePrompt, PRewrite, MORL-Prompt\ \cite{deng2022rlprompt,kwon2024stableprompt,kong2024prewrite,jafari2024morl}). StablePrompt, for instance, adopts an adaptive PPO variant because vanilla PPO collapses on discrete prompt tokens. Reflection-based methods (ProTeGi~\cite{pryzant2023protegi}, TextGrad~\cite{yuksekgonul2024textgrad}, PromptAgent~\cite{wang2024a}, GEPA~\cite{gepa}) instead use an LLM to read execution traces and propose edits, often with UCB, beam/tree search, or Pareto selection.

\textbf{Compound AI systems and cost controls.} Compound systems chain language modules under shared control flow. DSPy~\cite{dspy} treats the program as a compiler target for joint prompt optimization; MIPROv2~\cite{opsahl2024mipro} adds Bayesian prompt search and GEPA~\cite{gepa} reflective rewriting with Pareto merging, often at thousands of rollouts. In parallel, RLHF/RLVR analyses suggest some gains credited to policy updates are reachable by sampling and selection: Yue et al.~\cite{yue2025rlvr} use pass@$k$ to argue RLVR mainly sharpens the base model's sampling distribution, and RAFT~\cite{dong2023raft} shows best-of-$N$ reward-ranked fine-tuning approaches PPO-style alignment. DCPS brings the same control to prompt optimization, with RCEI making marginal returns per token explicit.

\section{Demonstration-Conditioned Prompt Search}

\subsection{Task Definition}
Following the definition of compound AI systems~\cite{dspy,gepa}, we write the target system as $\Phi=(M,C,\mathcal{X},\mathcal{Y})$, where $M=\langle M_1,\ldots,M_{|M|}\rangle$ is a sequence of language modules, $C$ is the control flow, and $\mathcal{X},\mathcal{Y}$ are the global input and output schemas. Each module $M_i=(\pi_i,\theta_i,\mathcal{X}_i,\mathcal{Y}_i)$ contains a prompt $\pi_i$, fixed parameters $\theta_i$, and local schemas. Let $\Pi_\Phi$ collect all module prompts and $\Theta_\Phi$ the fixed target-model weights. Given a task distribution $\mathcal{T}$ over instances $(x,m)$ and a metric $\mu\in[0,1]$, prompt optimization seeks, under rollout budget $B$:
\begin{equation}
\Pi^*_\Phi = \arg\max_{\Pi_\Phi}\; \mathbb{E}_{(x,m)\sim\mathcal{T}}\left[\mu(\Phi(x;\Pi_\Phi,\Theta_\Phi),m)\right]
\end{equation}
A rollout is one evaluation call to $\Phi$ under a candidate prompt configuration. This covers both single-prompt benchmarks ($|M|=1$) and compound pipelines (where $\Pi_\Phi$ is jointly optimized across modules).

\subsection{DCPS Pipeline}
\label{sec:pipeline}

DCPS uses one optimization signal: the validation performance of prompts generated from randomly sampled demonstrations. The demonstrations shape the generator's proposal but do not by themselves become part of the final inference prompt. Given a target system $\Phi$, a training set $D_{\mathrm{train}}$, a fixed validation subset $D_{\mathrm{val}}^{\mathrm{sub}}$, a frozen generator $p_{\theta_G^{(0)}}$, and metric $\mu$, DCPS iterates four stages on the candidate prompt set:
\begin{itemize}
    \item \textbf{Demonstration sampling $\mathcal{S}$.} Sample $k$ examples at random from $D_{\mathrm{train}}$ and format them into a meta-prompt $\mathcal{I}^{(t)}$ for the generator.
    \item \textbf{Candidate generation $\mathcal{G}_{\theta_G^{(0)}}$.} Draw candidate prompts $\Pi^{(t)}\sim p_{\theta_G^{(0)}}(\cdot\mid\mathcal{I}^{(t)})$ from the frozen generator, with no parameter update, trace-based reflection, or rewrite loop.
    \item \textbf{Evaluation $\mathcal{E}$.} Score each generated prompt on the fixed validation subset $D_{\mathrm{val}}^{\mathrm{sub}}$ under $\mu$.
    \item \textbf{Selection $\mathcal{F}$.} Retain the best candidate(s): greedy replacement in compound runs, or Top-$K$ retention in StablePrompt-DCPS.
\end{itemize}
The resulting control operator is
\begin{equation}
\mathcal{O}_{\mathrm{DCPS}}(\mathcal{T},\Phi,B) = \mathcal{F}\circ\mathcal{E}\circ\mathcal{G}_{\theta_G^{(0)}}\circ\mathcal{S}.
\end{equation}

At evaluation time the target system receives the generated prompt $\Pi^{(t)}$; the sampled demonstrations affect it only through the generator's meta-prompt, unless the generator copies them into its output. This separates DCPS from few-shot example selection and from few-shot in-context evaluation, while making the borrowed component explicit: APE-style demonstration-conditioned proposal coupled to score-based selection~\cite{zhou2023ape}, without policy updates, trace reflection, prompt mutation, or merge operations.

\subsection{Two DCPS Audits}
\label{sec:two-instantiations}
We instantiate the DCPS control template in two settings, comparing against StablePrompt, GEPA, and MIPROv2-Heavy (the heavy-budget MIPROv2 configuration) as three representative sources of complexity in modern APO: policy-gradient updates, trace reflection with prompt evolution, and Bayesian search over instruction--demonstration programs. The selector is matched to each host: StablePrompt-DCPS keeps StablePrompt's Top-$K$ queue and training-batch rewards, while DCPS-Compound uses greedy validation replacement. These serve as a \emph{policy-update audit} of StablePrompt and a \emph{compound-system audit} against GEPA-style reflection and MIPROv2-style Bayesian search (Algorithms~\ref{alg:no_update} and~\ref{alg:dcps}).

\subsubsection{StablePrompt-DCPS audit.}
StablePrompt~\cite{kwon2024stableprompt} already contains the proposal stage DCPS isolates: an agent generator receives randomly sampled examples in a meta-prompt, proposes candidates, and is then updated with PPO-style rewards. \textbf{StablePrompt-DCPS} (Algorithm~\ref{alg:no_update}) freezes the initial generator weights $\theta_G^{(0)}$ and removes PPO updates, while keeping random $k$-shot sampling, the original reward computation, and Top-$K$ retention. It also inherits StablePrompt's data protocol: demonstrations are drawn from $D_{\mathrm{val}}^{\mathrm{demo}}$ and candidates are scored on training batches, reversing the default DCPS data roles in \S\ref{sec:pipeline}. Applied identically to PPO and DCPS, this protocol leaves the comparison unaffected. The gap between StablePrompt-PPO and StablePrompt-DCPS estimates what the policy update adds beyond proposal and selection.

\subsubsection{DCPS-Compound for compound AI systems.}
For compound AI systems, DCPS-Compound follows the task and rollout protocol used for GEPA-family and MIPROv2-Heavy optimizers~\cite{gepa,opsahl2024mipro} (Algorithm~\ref{alg:dcps}), auditing what trace reflection, Pareto selection, and Bayesian instruction--demonstration search add beyond the generate-and-select control. Each iteration samples demonstrations from $D_{\mathrm{train}}$, queries the frozen generator, evaluates on $D_{\mathrm{val}}^{\mathrm{sub}}$, and replaces the incumbent when validation improves.

\begin{figure}[t]
\centering
\begin{minipage}[t]{0.48\textwidth}
\refstepcounter{algorithm}
\label{alg:no_update}
\hrule\smallskip
{\footnotesize\textbf{Algorithm~\thealgorithm} StablePrompt-DCPS}
\smallskip\hrule\smallskip
{\footnotesize
\begin{algorithmic}[1]
\REQUIRE Training set $D_{\mathrm{train}}$, validation set $D_{\mathrm{val}}^{\mathrm{demo}}$, frozen generator $p_{\theta_G^{(0)}}$, target system $\Phi$, metric $\mu$, epochs $T$, candidates per batch $N$, few-shot size $k$, queue capacity $K$
\STATE $\mathcal{P}\leftarrow\emptyset,\; t\leftarrow 0$
\FOR{$e=1$ to $T$}
    \FOR{$B_j \in \mathrm{Batches}(D_{\mathrm{train}})$}
        \STATE $t\leftarrow t+1$
        \STATE $E^{(t)}\leftarrow\mathrm{Sample}(D_{\mathrm{val}}^{\mathrm{demo}},k)$ \COMMENT{$\mathcal{S}$}
        \STATE $\mathcal{I}^{(t)}\leftarrow \mathrm{FmtMeta}(E^{(t)})$
        \STATE $\{\Pi_i^{(t)}\}_{i=1}^{N}\sim p_{\theta_G^{(0)}}(\cdot\mid\mathcal{I}^{(t)})$ \COMMENT{$\mathcal{G}$}
        \FOR{$i=1$ to $N$}
            \STATE $s_i^{(t)}\leftarrow\mathrm{Eval}(\Pi_i^{(t)}, B_j)$ \COMMENT{$\mathcal{E}$}
            \STATE $c_i^{(t)}\leftarrow(\Pi_i^{(t)},s_i^{(t)})$
            \STATE $\mathcal{P}\leftarrow\mathrm{TopK}(\mathcal{P}\cup\{c_i^{(t)}\}, K)$ \COMMENT{$\mathcal{F}$}
        \ENDFOR
    \ENDFOR
\ENDFOR
\RETURN $\mathcal{P}$
\end{algorithmic}
}
\smallskip\hrule
\end{minipage}%
\hfill
\begin{minipage}[t]{0.48\textwidth}
\refstepcounter{algorithm}
\label{alg:dcps}
\hrule\smallskip
{\footnotesize\textbf{Algorithm~\thealgorithm} DCPS-Compound}
\smallskip\hrule\smallskip
{\footnotesize
\begin{algorithmic}[1]
\REQUIRE Training set $D_{\mathrm{train}}$, validation set $D_{\mathrm{val}}$, frozen generator $p_{\theta_G^{(0)}}$, target system $\Phi$, metric $\mu$, iterations $I$, few-shot size $k$, val.~subset $n_v$
\STATE $\Pi^*\leftarrow\mathrm{null},\; s^*\leftarrow -\infty$;\; $D_{\mathrm{val}}^{\mathrm{sub}}\leftarrow\mathrm{Sample}(D_{\mathrm{val}},n_v)$
\FOR{$t=1$ to $I$}
    \STATE $E^{(t)}\leftarrow\mathrm{Sample}(D_{\mathrm{train}},k)$ \COMMENT{$\mathcal{S}$}
    \STATE $\mathcal{I}^{(t)}\leftarrow \mathrm{FmtMeta}(E^{(t)})$
    \STATE $\Pi^{(t)}\sim p_{\theta_G^{(0)}}(\cdot\mid\mathcal{I}^{(t)})$ \COMMENT{$\mathcal{G}$}
    \STATE $s^{(t)}\leftarrow\mathrm{Eval}(\Pi^{(t)}, D_{\mathrm{val}}^{\mathrm{sub}})$ \COMMENT{$\mathcal{E}$}
    \IF{$s^{(t)} > s^*$}
        \STATE $\Pi^*\leftarrow \Pi^{(t)};\; s^*\leftarrow s^{(t)}$ \COMMENT{$\mathcal{F}$}
    \ENDIF
\ENDFOR
\RETURN $\Pi^*$
\end{algorithmic}
}
\smallskip\hrule
\end{minipage}
\caption{Left: StablePrompt-DCPS. Right: DCPS-Compound. $\mathrm{FmtMeta}(E)$ formats the $k$-shot examples into a meta-prompt for the generator; $\mathrm{Eval}(\Pi,D)=\frac{1}{|D|}\sum_{(x,m)\in D}\mu(\Phi(x;\Pi,\Theta_\Phi),m)$. In StablePrompt-DCPS, $D_{\mathrm{val}}^{\mathrm{demo}}$ supplies only demonstrations and rewards are computed on training batches $B_j$.}
\end{figure}

\section{Experimental Setup}
\label{sec:exp-setup}

We evaluate StablePrompt-DCPS on the original StablePrompt benchmarks and DCPS-Compound against GEPA, GEPA-MERGE, and MIPROv2-Heavy on compound systems. The runs reuse the original implementations and published defaults; costs are measured from WandB summaries, local logs, and API-provider records.

\subsection{StablePrompt-DCPS Ablation}
\label{sec:phase1-setup}
We follow Kwon et al.'s full StablePrompt configuration~\cite{kwon2024stableprompt}. Both the target and agent models are \texttt{gemma-1.1-7B-it}~\cite{gemma2024}. We evaluate this setup on the same three task families with the original scoring protocols: (1) \textbf{text classification} (6 GLUE/SuperGLUE subsets~\cite{wang2018glue,wang2019superglue}), (2) \textbf{multi-task QA} (57-subject MMLU~\cite{hendrycks2020mmlu}), and (3) \textbf{instruction induction \& generation} (24 Instruction Induction (II)~\cite{honovich2022instruction} + 18 Big-Bench Instruction Induction (BBII)~\cite{zhou2023ape} subsets, split into BBII-TC and BBII-Gen). Removing gradient computation and PPO optimizer states lets DCPS run inference-only on one NVIDIA A40 48\,GB, whereas StablePrompt-PPO follows the original A100 80\,GB setup. In our reproduction, we fix a train/test overlap in II and missing batch weights in StablePrompt's Softmax-difference metric. Following StablePrompt, we report the mean over three seeds of the best test score among the Top-5 prompts ranked by training reward; this protocol is optimistic in absolute terms but is applied identically to PPO and DCPS, so the PPO--DCPS comparison is unaffected.

\subsection{Compound AI System Optimization Setup}
\label{sec:phase2-setup}
We evaluate two backbones, \texttt{Qwen3-8B}~\cite{qwen32025} and \texttt{GPT-4.1-mini}~\cite{openai2025gpt41}, using each as both the target system and the prompt generator. The six benchmarks cover multi-step retrieval tasks, HotpotQA~\cite{yang2018hotpotqa} and HoVer~\cite{jiang2020hover}, plus single- or two-stage reasoning and control tasks: IFBench (verifiable instruction following)~\cite{pyatkin2025ifbench}, PUPA (PAPILLON privacy delegation)~\cite{li2025papillon}, AIME-2025~\cite{aime2025}, and LiveBench-Math~\cite{livebench}. DCPS-Compound uses task-specific rollout budgets and validation subset sizes $n_v\!\in\![15,50]$; for each replicate, the reported prompt ranks first by validation score, and when multiple logged replicates are available, we select the displayed result using validation performance only, never test scores.

\subsection{Cost and Efficiency Metrics}
\label{sec:rcei}
We report accuracy together with rollout budgets and measured cost. A \emph{rollout} is one target-system evaluation or generator call, so the per-task DCPS-Compound budget has the closed form $\mathrm{rollouts}=(n_v+1)\times I$, where $n_v$ is the validation-subset size and $I$ the number of iterations. This reproduces every rollout budget in Table~\ref{tab:compound_results}(b) (AIME-2025 $320$, IFBench/LiveBench-Math $620$, PUPA $920$, HotpotQA $1{,}020$, HoVer $7{,}068$; macro $1{,}761$). Validation dominates the count (e.g.\ $50$ of every $51$ HotpotQA rollouts), so DCPS's cost is due almost entirely to proposal and validation. For cost-normalized comparison, we use relative error reduction (RER), which normalizes gains by the remaining headroom over the unoptimized prompt:
\begin{equation}
\mathrm{RER}(\mathcal{M}\parallel\mathcal{B}) = \frac{\mathcal{M}-\mathcal{B}}{100\%-\mathcal{B}}\times 100\%,
\end{equation}
where $\mathcal{B}$ is the unoptimized prompt score and $\mathcal{M}$ is the score achieved by the optimized method. Cost is reported in millions of price-weighted tokens across optimization and final test evaluation:
\begin{equation}
T = T_{\text{optimizer}} + T_{\text{test}}, \qquad T_{(\cdot)}=T^{(\cdot)}_{\text{in}} + 4\,T^{(\cdot)}_{\text{out}},
\end{equation}
using the approximate GPT-4.1-mini output-to-input price ratio. RCEI reports relative error reduction per log cost multiplier:
\begin{equation}
\mathrm{RCEI}(\mathcal{M}\parallel\mathcal{B}) = \frac{\mathrm{RER}(\mathcal{M}\parallel\mathcal{B})}{\log_{2}\left(T_{\mathcal{M}}/T_{\mathcal{B}}\right)}.
\end{equation}
Since optimization always consumes tokens, $T_{\mathcal{M}}>T_{\mathcal{B}}$ and the denominator is strictly positive. RCEI is an auxiliary view, reported with accuracy, rollout counts, and dollar costs; macro RER and RCEI are arithmetic means over per-task values.

\section{Main Results and Diagnostic Findings}
\label{sec:main_results_diagnostic}

We report accuracy alongside rollout counts and costs for both audits. \S\ref{sec:gemma-results} uses StablePrompt-DCPS to ask what PPO adds over the frozen-generator control, and \S\ref{sec:phase2-results} asks the same of reflection and Bayesian search. The two differ in granularity: the StablePrompt study is a single-variable ablation, while the compound-system study is a control-based cost--performance comparison, not a budget-matched ranking.

\subsection{StablePrompt-PPO vs StablePrompt-DCPS}
\label{sec:gemma-results}

\begin{figure}[tb]
\centering
\includegraphics[width=0.62\textwidth]{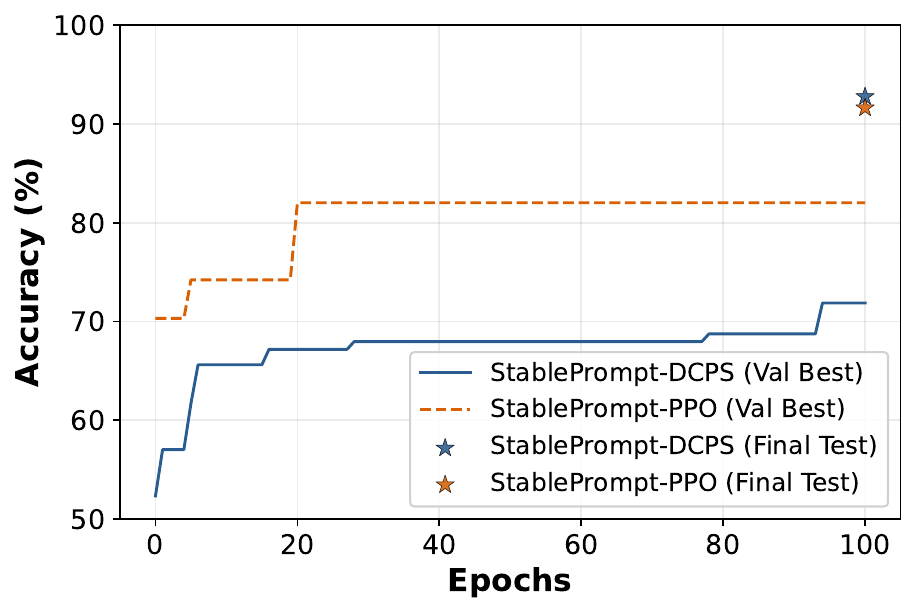}
\caption{Running-best SST-2 validation accuracy for StablePrompt-DCPS vs.\ PPO on \texttt{gemma-1.1-7B-it}; stars mark the final test score at epoch 100.}
\label{fig:trajectories}
\end{figure}

\begin{table}[!ht]
\centering
\caption{StablePrompt-PPO vs.\ StablePrompt-DCPS on \texttt{gemma-1.1-7B-it}. Scores are macro-averages over subsets; $\Delta$ (PPO$-$DCPS), 95\% CI, and Cohen's $d$ are computed from per-subset differences (vs.\ PPO@30 epochs for extended II runs). Times and measured GPU costs are reported per run.}
\label{tab:gemma_ablation}
{\setlength{\tabcolsep}{1.5pt}
\resizebox{\textwidth}{!}{
\begin{tabular}{@{}lcccccccl@{}}
\toprule
\textbf{Task Family} & \textbf{PPO} & \textbf{DCPS} & $\boldsymbol{\Delta}$ & \textbf{95\% CI} & $\boldsymbol{d}$ & \textbf{DCPS (h/\$)} & \textbf{PPO (h/\$)} & \textbf{Note} \\ \midrule
GLUE/SuperGLUE & 76.7 & \textbf{77.1} & $-0.4$ & $[-2.30,+1.97]$ & $-0.14$ & 4.83/1.93 & 3.97/5.96 & comparable \\
BBII-TC        & \textbf{57.6} & 57.2 & $+0.4$ & $[-0.68,+1.63]$ & $+0.20$ & 3.38/1.35 & 3.17/4.76 & comparable \\
BBII-Gen       & \textbf{63.0} & 60.7 & $+2.3$ & $[+0.15,+4.35]$ & $+0.79$ & 2.66/1.06 & 1.84/2.76 & PPO better \\
II @30 epochs  & \textbf{48.7} & 44.3 & $+4.4$ & $[+0.15,+8.70]$ & $+0.41$ & 2.76/1.10 & 6.65/9.98 & PPO better \\
II @100 epochs & --   & 48.6 & $+0.1$ & $[-2.67,+2.63]$ & $+0.01$ & 14.74/5.90 & -- & gap closed \\
MMLU           & \textbf{55.9} & 54.1 & $+1.8$ & $[+0.99,+2.74]$ & $+0.53$ & 6.58/2.63 & 6.38/9.57 & PPO better \\
\bottomrule
\end{tabular}}}
\end{table}

Table~\ref{tab:gemma_ablation} shows PPO updates have limited marginal value on classification: on GLUE/SuperGLUE~\cite{wang2018glue,wang2019superglue} and BBII-TC~\cite{zhou2023ape}, DCPS and PPO do not differ significantly ($|\Delta|\le0.4$~pp), and Figure~\ref{fig:trajectories} shows DCPS tracking PPO throughout SST-2 training. The picture changes where behavior is less template-like: PPO leads by $+2.3$~pp on BBII-Gen~\cite{zhou2023ape} and $+1.8$~pp on MMLU~\cite{hendrycks2020mmlu}, so policy updates still help when the target is hard to reach by random conditioning. On II~\cite{honovich2022instruction}, extending DCPS to 100 epochs closes the gap with PPO at 30 epochs ($+0.1$~pp). Because PPO is not re-run at matching epochs, this result suggests that extra search budget can substitute for the policy update; it does not show that DCPS strictly dominates. Where accuracy is matched, removing the policy update eliminates gradient computation and optimizer state, enabling inference-only runs on a cheaper GPU tier: measured single-run cost falls from \$4.76--\$5.96 (A100) to \$1.35--\$1.93 (A40), about one-third as much under our rates.

\subsection{DCPS on Compound AI Systems}
\label{sec:phase2-results}

Table~\ref{tab:compound_results}(a) summarizes the cost--performance trade-off. With less than half of GEPA's average rollout budget ($1{,}761$ vs.\ $3{,}936$) and a lower API cost, DCPS-Compound attains both the highest macro RER ($+26.23\%$) and the highest macro RCEI among optimized methods, while matching GEPA's absolute macro accuracy on GPT-4.1-mini ($61.14$, single seed; see \S\ref{sec:discussion-limitations}). MIPROv2-Heavy~\cite{opsahl2024mipro}, run at GEPA's full budget, trails on both relative error reduction and cost-efficiency.

The main table reports one selected rollout budget per method--task cell: GEPA-family methods at their published defaults and DCPS-Compound at the budgets in Table~\ref{tab:compound_results}. For HoVer, a small-budget DCPS run matching the other tasks was weak, so the table reports the GEPA-aligned extended run. \S\ref{sec:budget-matched} reports a budget-sensitivity analysis on three tasks, leaving a full sweep for future work.

\begin{table}[!t]
\centering
\caption{Compound-system results. \textbf{(a)} Cost--performance summary on GPT-4.1-mini, macro-averaged across the six benchmarks ($\overline{\log_2(T_\mathcal{M}/T_\mathcal{B})}$ with $T$ as in \S\ref{sec:rcei}). \textbf{(b)} Task-level accuracy on Qwen3-8B and GPT-4.1-mini; macro-average is the unweighted mean over six benchmarks, and the bottom block lists per-task rollout budgets (shared across backbones). In (b), LB-Math\,=\,LiveBench-Math and AIME\,=\,AIME-2025.}
\label{tab:compound_results}

{\footnotesize\textbf{(a) Cost--performance summary (GPT-4.1-mini, macro-averaged)}}

\smallskip
\resizebox{\textwidth}{!}{
\begin{tabular}{@{}lcccccc@{}}
\toprule
\textbf{Method ($\mathcal{M}$)} & \textbf{RER} & \textbf{$T_\mathcal{M}$} & \textbf{$\overline{\log_2(T_\mathcal{M}/T_\mathcal{B})}$} & \textbf{RCEI} & \textbf{Rollouts} & \textbf{Cost} \\ \midrule
Baseline (unoptimized prompt, $\mathcal{B}$) & $0$ & $1.06$ & $0$ & -- & -- & \$0.43 \\
MIPROv2-Heavy & $+12.70\%$ & $27.11$ & $4.67$ & $+2.66$ & $3{,}936$ & \$10.84 \\
GEPA          & $+24.15\%$ & $28.36$ & $4.73$ & $+5.30$ & $3{,}936$ & \$11.34 \\
GEPA-MERGE    & $+24.87\%$ & $24.07$ & $4.50$ & $+5.90$ & $3{,}936$ & \$9.63 \\
DCPS-Compound & $\mathbf{+26.23\%}$ & $\mathbf{19.26}$ & $\mathbf{3.85}$ & $\mathbf{+7.20}$ & $\mathbf{1{,}761}$ & $\mathbf{\$7.70}$ \\
\bottomrule
\end{tabular}}

\smallskip
{\footnotesize\textbf{(b) Task-level accuracy}}

\smallskip
{\setlength{\tabcolsep}{2pt}
\resizebox{\textwidth}{!}{
\begin{tabular}{@{}llccccccc@{}}
\toprule
\textbf{Backbone} & \textbf{Method} & \textbf{Hotpot} & \textbf{IFBench} & \textbf{HoVer} & \textbf{PUPA} & \textbf{AIME} & \textbf{LB-Math} & \textbf{Macro} \\ \midrule
\multirow{5}{*}{Qwen3-8B}
 & Baseline       & 40.33 & 38.61 & 33.67 & 81.55 & 47.33 & 65.01 & 51.08 \\
 & MIPROv2-Heavy  & 58.67 & 38.61 & 44.00 & 78.67 & 58.00 & 67.46 & 57.57 \\
 & GEPA           & \textbf{63.33} & 40.65 & 53.33 & 87.57 & 59.33 & \textbf{70.57} & \textbf{62.46} \\
 & GEPA-MERGE     & 61.33 & 29.08 & 52.00 & 84.81 & \textbf{62.67} & 66.28 & 59.36 \\
 & DCPS-Compound  & 61.33 & \textbf{43.88} & \textbf{56.67} & \textbf{90.02} & 55.33 & 65.08 & 62.05 \\ \midrule
\multirow{5}{*}{GPT-4.1-mini}
 & Baseline       & 36.10 & 48.13 & 40.33 & 80.81 & 40.00 & 55.80 & 50.20 \\
 & MIPROv2-Heavy  & 55.00 & 51.19 & 47.33 & 83.59 & 46.67 & 57.30 & 56.85 \\
 & GEPA           & \textbf{65.00} & 49.83 & 49.67 & 90.10 & \textbf{48.00} & \textbf{64.21} & \textbf{61.14} \\
 & GEPA-MERGE     & 63.33 & 49.49 & 49.00 & 93.52 & 46.67 & 61.14 & 60.53 \\
 & DCPS-Compound  & 59.00 & \textbf{51.53} & \textbf{54.67} & \textbf{94.09} & \textbf{48.00} & 59.52 & \textbf{61.14} \\ \midrule
\multicolumn{9}{@{}l}{\emph{Total optimization budget (\# rollouts)}} \\
\multicolumn{2}{@{}l}{GEPA/GEPA-MERGE/MIPROv2-Heavy} & 6{,}871 & 3{,}593 & 7{,}051 & 2{,}426 & 1{,}839 & 1{,}839 & 3{,}936 \\
\multicolumn{2}{@{}l}{DCPS-Compound} & 1{,}020 & 620 & 7{,}068 & 920 & 320 & 620 & 1{,}761 \\
\bottomrule
\end{tabular}}}
\end{table}

Table~\ref{tab:compound_results}(b) reveals the underlying mechanism boundary. DCPS-Compound is strongest on single- or two-stage tasks that hinge on precise instruction following (IFBench, PUPA), and on HoVer once adequately budgeted (\S\ref{sec:budget-matched}). GEPA's advantages are concentrated on reasoning-heavy tasks: HotpotQA ($+6.00$~pp on GPT-4.1-mini), LiveBench-Math (both backbones), and AIME-2025 with Qwen3-8B. DCPS offsets these on IFBench, PUPA, and budgeted HoVer, so the GPT-4.1-mini macro tie reflects offsetting per-task results, not method equivalence. GEPA-MERGE, by contrast, falls below even the unoptimized Qwen3-8B IFBench baseline, indicating that crossover-based prompt merging can be unstable on strict-format tasks.

\paragraph{Prompt length and the shapes of optimized prompts.}
Beyond rollouts and dollars, the optimized prompts differ sharply in length (Table~\ref{tab:promptlen}). DCPS produces the shortest optimized prompt on every benchmark, measured in both characters and tokens counted with \texttt{tiktoken}'s \texttt{cl100k\_base} encoding. The ratios are not interchangeable: because DCPS's LiveBench-Math prompt is LaTeX-dense, it is $3.4\times$ shorter than GEPA's in characters but $2.9\times$ shorter in tokens; we therefore base cost claims on token counts. The gap is qualitative: DCPS specifies a compact \emph{output contract} while GEPA accumulates a long domain \emph{playbook}. On PUPA (GPT-4.1-mini), DCPS's 742-character redaction-module prompt states one transferable redaction rule and beats the baseline by $+13.3$~pp, whereas GEPA's $6{,}074$-character prompt hard-codes trace-distilled carve-outs that do not transfer to the held-out non-programming mix and scores below DCPS ($90.10$ vs.\ $94.09$). Where the reward is a transferable contract, the minimal prompt can generalize well; where GEPA leads, as on HotpotQA and LiveBench-Math, its longer trace-conditioned prompts are consistent with---but do not by themselves establish---a benefit from localized feedback. Verbatim prompts for all cells are in the released artifacts.

\begin{table}[!t]
\centering
\caption{Optimized-prompt length in characters~/~tokens (GPT-4.1-mini), with tokens counted using \texttt{tiktoken}'s \texttt{cl100k\_base} encoding. DCPS is shortest on every benchmark.}
\label{tab:promptlen}
\resizebox{\textwidth}{!}{
\begin{tabular}{@{}lccc@{}}
\toprule
\textbf{Benchmark} & \textbf{DCPS (char / tok)} & \textbf{GEPA} & \textbf{MIPROv2-Heavy} \\ \midrule
AIME-2025      & \textbf{1{,}339 / 227} & 6{,}625 / 1{,}330 & 4{,}460 / 748 \\
LiveBench-Math & \textbf{1{,}536 / 337} & 5{,}295 / 984   & 5{,}519 / 950 \\
HoVer          & \textbf{2{,}471 / 460} & 12{,}135 / 2{,}309 & 16{,}401 / 2{,}922 \\
HotpotQA       & \textbf{3{,}009 / 591} & 10{,}479 / 2{,}105 & 14{,}681 / 2{,}521 \\
IFBench        & \textbf{1{,}458 / 259} & 6{,}598 / 1{,}254 & 11{,}027 / 1{,}893 \\
PUPA           & \textbf{1{,}459 / 270} & 6{,}074 / 1{,}180 & 11{,}021 / 1{,}978 \\
\bottomrule
\end{tabular}}
\end{table}

\subsection{Budget Sensitivity and the Generalization Gap}
\label{sec:budget-matched}
The per-task budgets in Table~\ref{tab:compound_results} raise a natural question: when does a larger rollout budget help, and when does it merely pressure the selector? We extend DCPS on three diagnostic tasks under the same validation-only selection rule, increasing the HotpotQA budget to $6{,}120$ rollouts, the HoVer budget toward GEPA scale, and the LiveBench-Math budget to $1{,}860$ rollouts. These runs are not new leaderboard entries; they test whether validation-selected prompts continue to generalize as the candidate pool grows. For HoVer, the main table reports the extended DCPS run because the initial small-budget run (620 rollouts, matching the other small-budget tasks) was clearly under-budgeted ($44.00$ on Qwen3-8B and $45.67$ on GPT-4.1-mini).

The probes separate three regimes. \textbf{HoVer} is mainly a coverage bottleneck: with GEPA-scale sampling, DCPS surpasses GEPA on both backbones while keeping validation--test gaps small ($\le3.3$~pp), so the initial GEPA advantage was a budget artifact. \textbf{HotpotQA} is different: extending DCPS to $6{,}120$ rollouts raises GPT-4.1-mini from $59.00$ to $63.67$ (narrowing the gap to GEPA's $65.00$ from $6.00$ to $1.33$~pp) but leaves Qwen3-8B essentially flat ($61.33\!\to\!60.33$, vs.\ GEPA's $63.33$). Extra rollouts thus narrow the gap on GPT-4.1-mini but do not remove GEPA's edge on either backbone. This pattern is consistent with trace-level feedback helping to localize failures across retrieval and reasoning steps, but the present comparison does not isolate trace feedback from GEPA's other search mechanisms.

\textbf{LiveBench-Math} is the cautionary case: more rollouts raise validation accuracy but lower held-out accuracy (gaps of $20.95$ and $13.50$~pp in single-seed results), so the rollout budget represents not only compute but also statistical pressure on the selector.

\paragraph{Failure modes: validation representativeness and demonstration sampling.}
Two factors govern when validation-only selection misleads (Table~\ref{tab:valgap}). First, \emph{subset representativeness} matters more than size: on AIME-2025 with Qwen3-8B (AIME-Qwen; $45$ validation items), a fixed head slice of $15$ inflates validation accuracy to $73.3$ but transfers to only $46.0$ on the test set, $\approx9$~pp worse than a seeded random sample of $15$ ($66.7\!\to\!55.3$) or the full set ($55.6\!\to\!53.3$); we use random sampling throughout. Second, the val$\to$test gap widens when the subset is small and validation accuracy saturates (IFBench, $\ge44$~pp) and shrinks as the subset grows (AIME-Qwen, $11.4$~pp)---a winner's-curse effect shared by any validation-selection method, including GEPA. Demonstration sampling (the $\mathcal{S}$ step) is a third lever: redrawing demonstrations introduces additional variance that the reported single-run points do not isolate; controlled noise injection and multi-seed analysis remain future work.

\begin{table}[htbp]
\centering
\caption{Validation$\to$test gap under validation-only selection. The gap is largest where the validation subset is small and validation accuracy saturates, and shrinks as the subset grows.}
\label{tab:valgap}
\resizebox{0.78\textwidth}{!}{
\begin{tabular}{@{}llccc@{}}
\toprule
\textbf{Task (backbone)} & \textbf{opt rollouts} & \textbf{val} & \textbf{test} & \textbf{gap} \\ \midrule
IFBench (GPT-4.1-mini) & 620 & 96.7 & 51.5 & 45.2 \\
IFBench (Qwen3-8B)     & 620 & 88.3 & 43.9 & 44.4 \\
AIME-2025 (GPT-4.1-mini) & 320 & 80.0 & 48.0 & 32.0 \\
AIME-2025 (Qwen3-8B)     & 320 & 66.7 & 55.3 & 11.4 \\
\bottomrule
\end{tabular}}
\end{table}

\section{Discussion}

\subsection{Applicability Boundaries from Optimization Signal to Pipeline Complexity}
\label{sec:discussion-scenarios}
The results show a task- and backbone-dependent pattern rather than a strict dichotomy. On tasks bottlenecked by output formatting or short-context instruction following, random demonstrations often give the frozen generator enough signal for proposal and selection to approach the heavier optimizers at lower cost. PPO and GEPA still retain advantages on some open-ended and reasoning-heavy cells, though the compound comparison does not isolate which of GEPA's reflection and search components produce those gains. The control likely does well in the first group because a strong instruction-tuned generator already carries prompt-writing priors; consistent with this hypothesis, heavier optimizers lead by larger margins on some Qwen3-8B cells and reasoning-heavy tasks. This remains a hypothesis, not a controlled test of generator strength. The large validation--test gaps on IFBench and AIME-2025 further show that the apparent strength of any validation-selected optimizer depends on subset representativeness and candidate-pool size. Heavier optimizers should therefore be evaluated against a simple control using held-out performance, not validation performance alone.

\subsection{Limitations and Empirical Boundaries}
\label{sec:discussion-limitations}
\textbf{Non-identical settings:} StablePrompt uses \texttt{gemma-1.1-7B-it}, whereas compound runs use Qwen3-8B and GPT-4.1-mini to match GEPA; the audits therefore examine each mechanism in its native setting. \textbf{Budget and replication:} the compound table reports one budget per method--task cell, and DCPS probes only three tasks. These validation-selected single runs establish cost--performance points, not a statistically resolved ranking. \textbf{Optimization capacity:} without trace rewriting or module-level attribution, DCPS cannot localize cascading errors; whether this explains its deficits requires component ablations. \textbf{Cost and coverage:} RCEI depends on token pricing and rental rates; code generation, summarization, tool use, and longer interactive tasks remain untested.

\section{Conclusion}
We asked how much performance attributed to complex prompt optimizers remains after reducing them to generation and validation-based selection. Under the evaluated settings, this control is cost-competitive on structured and instruction-following tasks, while PPO and GEPA retain advantages in some open-ended and reasoning-heavy settings. It should serve as a baseline, not replace heavier optimizers; evaluations should report validation--test gaps, rollout counts, token usage, dollar costs, and accuracy.

\bibliographystyle{splncs04}
\bibliography{references}

\end{document}